\pdfoutput=1

\documentclass[runningheads,a4paper]{llncs}

\usepackage{amssymb}
\usepackage{graphicx}

\usepackage{url}
\urldef{\mailsa}\path|chuanqi.tan@gmail.com|
\urldef{\mailsb}\path|fcsun@mail.tsinghua.edu.cn| 

\newcommand{\keywords}[1]{\par\addvspace\baselineskip
\noindent\keywordname\enspace\ignorespaces#1}

\usepackage{algorithm} 
\usepackage{algorithmic}
\usepackage{multirow}
\usepackage{amsmath}
\usepackage{xcolor}
\usepackage{subfigure}
\usepackage[misc]{ifsym}

\begin{document}

\mainmatter  

\title{Multimodal Classification with Deep Convolutional-Recurrent Neural Networks for Electroencephalography}

\titlerunning{Multimodal Classification with Deep CNN-RNN for EEG}

\author{Chuanqi Tan\textsuperscript{\dag} 
    \and Fuchun Sun\textsuperscript{(\Letter)} 
    \and Wenchang Zhang \\
    Jianhua Chen 
    \and Chunfang Liu}

\authorrunning{Chuanqi Tan et al.}

\institute{State Key Laboratory of Intelligent Technology and Systems
    \\Tsinghua National Laboratory for Information Science and Technology (TNList)
    \\Department of Computer Science and Technology
    \\Tsinghua University, Beijing, China
    \\ \textsuperscript{\dag}\mailsa, \textsuperscript{(\Letter)}\mailsb}

\toctitle{Lecture Notes in Computer Science}
\maketitle

\begin{abstract}

Electroencephalography (EEG) has become the most significant input signal for brain computer interface (BCI) based systems. However, it is very difficult to obtain satisfactory classification accuracy due to traditional methods can not fully exploit multimodal information.
Herein, we propose a novel approach to modeling cognitive events from EEG data by reducing it to a video classification problem, which is designed to preserve the multimodal information of EEG. 
In addition, optical flow is introduced to represent the variant information of EEG. 
We train a deep neural network (DNN) with convolutional neural network (CNN) and recurrent neural network (RNN) for the EEG classification task by using EEG video and optical flow.
The experiments demonstrate that our approach has many advantages, such as more robustness and more accuracy in EEG classification tasks.
According to our approach, we designed a mixed BCI-based rehabilitation support system to help stroke patients perform some basic operations.

\end{abstract}

\section{Introduction}

For patients suffering from stroke, it is very meaningful to provide a communication method to deliver brain messages and commands to the external world apart from the normal nerve-muscle output pathway.
Due to natural and non-intrusive characteristics, most BCI systems select the EEG signal as input \cite{amiri2013review}. The biggest challenge in BCI is EEG classification, aiming to translate raw EEG signal into the commands of the human brain. This can be used to control external equipment, such as rehabilitation devices and other devices, when the EEG signal is decoded correctly.
However, traditional EEG classification methods can not obtain satisfactory result, one of the reasons is that some useful information has been ignored.
Deep learning, as a new classification platform, has recently received increased attention from researchers \cite{lecun2015deep,min2016deep}. It has been successfully applied to many classification problems, such as image classification \cite{schmidhuber2015deep}, video classification \cite{Ng2015Beyond} and speech recognition \cite{yu2014automatic}. However, deep learning has not been fully explored in EEG classification. 
Similar to the structure of the human brain, deep learning is particularly suitable for classification problems from which it is hard to extract hand-designed features. Therefore, deep learning has very promising prospects in the EEG classification field.

The contributions of this paper are as follows. Firstly, our approach reduces the EEG classification problem to a video classification problem, which is designed to utilize multimodal information. Secondly, optical flow has been introduced into this field to characterize the variant of EEG signal in the temporal dimension. Thirdly, a deep CNN-RNN network has been constructed, which is designed for EEG videos and optical flow. Finally, a mixed BCI-based rehabilitation support system is built using our approach.

The rest of this paper is organized as follows. Firstly, we will review related works in Section 2. Secondly, the method we proposed will be described in Section 3. Third, findings of our experiments will be presented in Section 4. Finally, conclusions and further steps will be discussed in Section 5.

\section{Related work}

In order to improve the accuracy of EEG classification, a lot of work has been carried out. The performance of this pattern recognition like system depends on both the features selected and the classification algorithms employed. Traditionally, a great variety of hand-designed features have been proposed such as band powers (BP) \cite{kaiser2011first}, power spectral density (PSD) values \cite{waldert2009review} and so on. In recent years, the common spatial pattern (CSP) \cite{ramoser2000optimal} has been proved to be an expressive feature of EEG signal. A lot of related work has been proposed such as CSSP, WCSP and SCSSP \cite{aghaei2016separable}. Unlike these single modal approaches, there are many researchers focusing on how to extract multimodal information from the EEG signal \cite{verma2014multimodal,Bashivan2015Learning} and how to fuse this information \cite{tan2017spatial}.

From hand-designed to data-driven features, deep learning has played a significant role in diverse fields where the artificial intelligence (AI) community has struggled for many years. 
Certainly, bioinformatics can also benefit from deep learning. In recent years, many public reviews \cite{Mamoshina2016Applications,Greenspan2016Guest} have been proposed to discuss deep learning applications in bioinformatics research. For example, \cite{An2014A} applying deep belief networks (DBN) to the frequency components of EEG signal to classify left-hand and right-hand motor imagery skills. \cite{Cecotti2011Convolutional} used CNN to decode P300 patterns, and \cite{Stober2014Using} used CNN to recognize rhythm stimuli. \cite{Soleymani2014Continuous} conducted an emotion detection and facial expressions study with both EEG signal and face images by RNN.

\section{Method}

\subsection{Preprocessing}

We are only interested in certain brain activities, and these signals need to be separated from background noise, and unnecessary artifacts must be eliminated.
In the preprocessing phase, we first apply the Butterworth filter with 0.5-50Hz as a bandpass filter to remove high-order noise in the signal. Then, a denoise Autoencoder (DAE) \cite{li2015feature} as a symmetrical neural network is used to denoise in an unsupervised manner. 
It is trained to rebuild the input to construct a robust feature representation. Autoencoders, like the principal components analysis (PCA), are usually trained to perform dimension reduction tasks, but the DAE is more useful in learning sparse representations of input. This means that a high-dimensional original signal can be represented by using a few representative atoms on a low-dimensional manifold, which is similar to sparse coding.

\subsection{EEG videos and optical flow}

Similar to speech signal, the most notable features of EEG signal reside in the frequency dimension, which is usually studied using a spectrogram of the signal. The feature vector formed by aggregating spectral measurements of all electrodes is the traditional method in EEG data analysis. However, these methods clearly ignore the locations of electrodes and the inherent information in spatial dimension. 
In our approach, for representing multimodal information, we propose to preserve the spatial structure by EEG image, apply frequency filters to represent the spectral dimension, and utilize the EEG videos to account for temporal evolutions in brain activity.

\begin{figure}
    \centering
    \subfigure[3D locations of electrodes]{
        \includegraphics[width=1.9in]{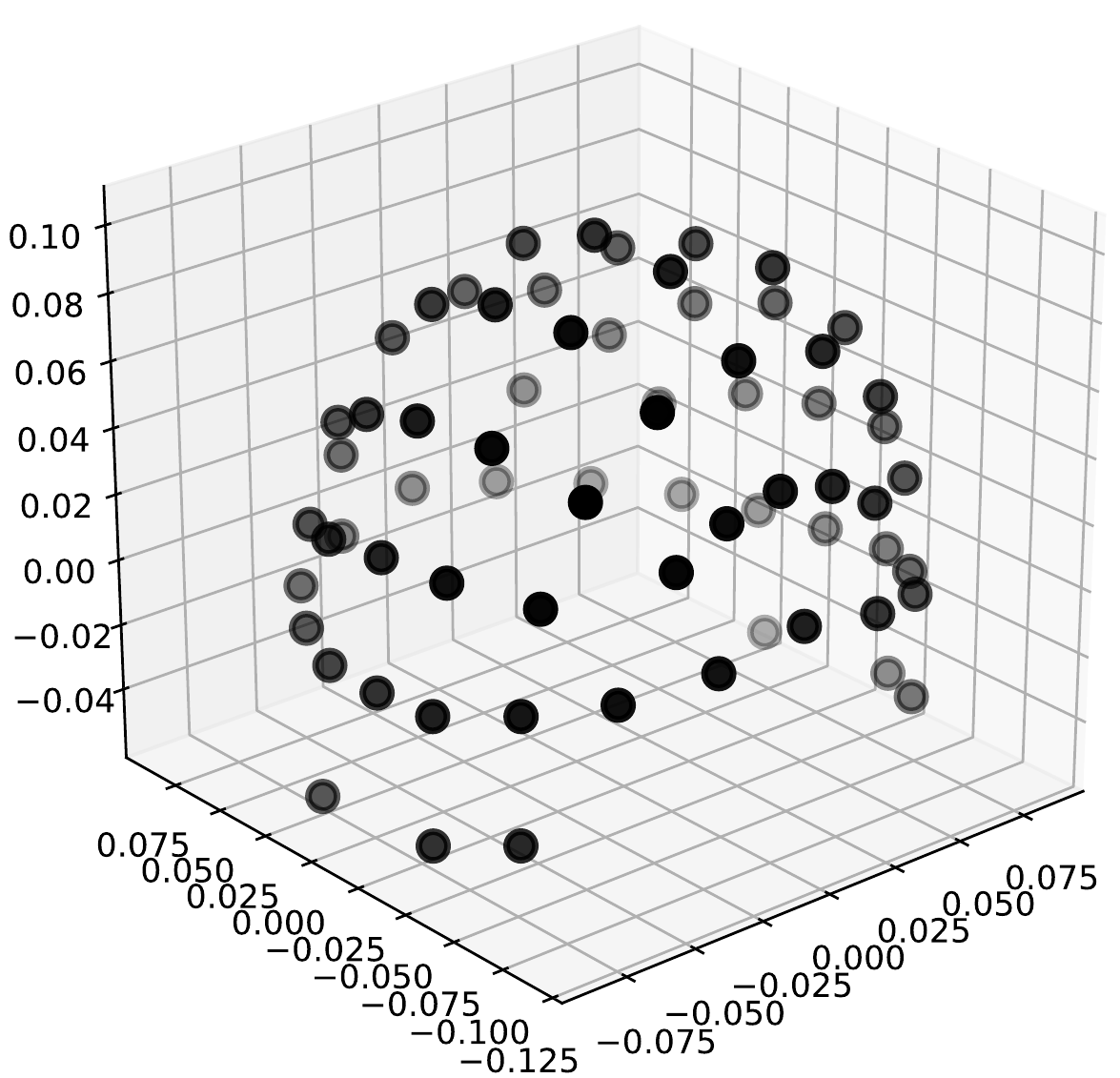}
        \label{3d location of electrodes}
    }
    \vspace{0in} \hspace{0in}
    \subfigure[Frames of EEG video]{
        \includegraphics[width=2.6in]{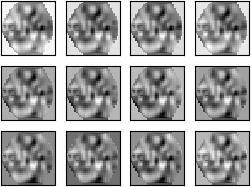}
        \label{generated eeg images}
    }
    \caption{Frames of EEG video generated from EEG signal by project the 3D locations of electrodes to 2D points via AEP algorithm}
    \label{3d location and EEG frame}
\end{figure}

Firstly, filtering is performed by using five frequency filters ($\alpha$: 8-13Hz, $\beta$: 14-30Hz, $\gamma$: 31-51 Hz, $\delta$: 0.5-3 Hz, $\theta$: 4-7Hz) to represent different EEG signal rhythms which correspond to different brain activity. According to the frequency characteristics of the EEG signal, we produced five different EEG dataset by these filters. 
Secondly, EEG images are generated for each EEG frame in time dimension. We project the 3D locations of electrodes (shown in Fig. \ref{3d location of electrodes}, unit of percentage) to 2D points by azimuthal equidistant projection (AEP) which borrows from mapping applications, and interpolate them to a 32*32 gray image. We refer to the collection of these EEG images on the time-line as EEG video. 
Compare to the EEG topographic maps used for EEG visualization, EEG images generated by AEP can maintain the distance between electrodes more accurately, which reflect more useful information in spatial dimension.
Finally, we split each EEG video into 12 segments and perform average operation in each segment. In this way, each EEG video is compressed into a 12-frame short video. The frames of a sample EEG video are shown in Fig. \ref{generated eeg images}.

Reducing the EEG classification problem to a video classification problem brings many benefits. The spatial structure of the electrodes has been preserved clearly. Many of the video classification techniques can also be applied to EEG signal. Due to the inherent structure of CNNs, it is more suited to image and video data classification. Moreover, there are many excellent CNNs such as AlexNet and GoogLeNet that can be used for EEG videos.

Optical flow \cite{Farneb2003Two} has been introduced by our approach to represent the variant information of EEG signal. Optical flow is widely used in most video classification method, because it can describe the obvious motion of objects in a visual scene by calculate the motion between two image frames which are taken at times $t$ and $t+\Delta t$ at every pixel position. Consider $f(x,y,t)$ is the pixel of location $(x,y)$ at time $t$, it moves by distance $(\Delta x,\Delta y)$ in next frame taken at $x+\Delta t$. These pixels has the same value, and the following brightness constancy constraint can be given:

\begin{equation}
f(x,y,t) = f(x+\Delta x, y+\Delta y, t+\Delta t)
\end{equation}

Assuming the movement to be small, take Taylor series approximation of right-hand side and ignoring higher-order terms in the Taylor series, we can get the following equation:

\begin{equation}
f(x+\Delta x, y+\Delta y, t+\Delta t) = f(x,y,t) + \frac{\partial f}{\partial x} \Delta x + \frac{\partial f}{\partial y} \Delta y + \frac{\partial f}{\partial t} \Delta t + \dots
\end{equation}

Then remove common terms and divide by $\Delta t$ to get:

\begin{equation}
\frac{\partial f}{\partial x} \frac{\Delta x}{\Delta t} + \frac{\partial f}{\partial y} \frac{\Delta y}{\Delta t} + \frac{\partial f}{\partial t} \frac{\Delta t}{\Delta t} 
= \frac{\partial f}{\partial x} u + \frac{\partial f}{\partial y} v + \frac{\partial f}{\partial t} 
= 0
\end{equation}

where $u=\Delta x / \Delta t$ and $v=\Delta y / \Delta t$. In this equation, $(u, v)$ is the value of optical flow at $f(x,y,t)$ which are responding to magnitude and direction respectively.

\begin{figure}[thpb]
	\centering
	\includegraphics[width=2.6in]{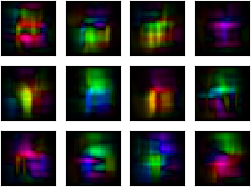}
	\caption{Visualization of optical flow extracted from EEG video}
	\label{optical flow images}
\end{figure} 

To utilize existing implementations and networks used for frame of EEG video, we store optical flow as an image and rescale it to a [0,255] range, and the visualization images are shown in Fig. \ref{optical flow images} by mapping direction to Hue value and mapping magnitude to Value plane on HSV image. In this way, optical flow can be processing using the same way as EEG image to learn the global description of EEG videos.

\subsection{Network architecture}

\begin{figure}[thpb]
    \centering
    \includegraphics[width=4.8in]{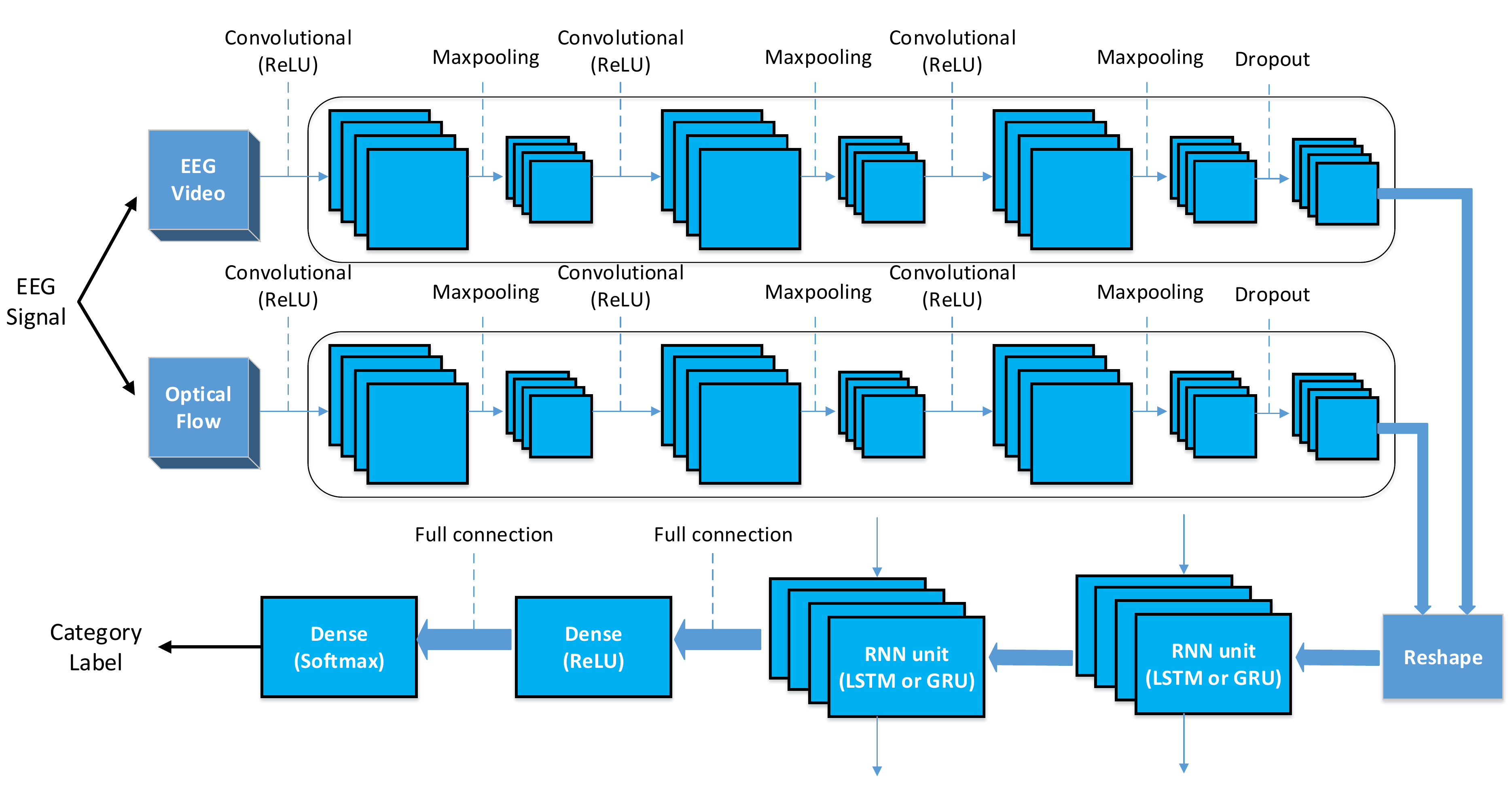}
    \caption{Architecture of our deep CNN-RNN network}
    \label{network architecture}
\end{figure}

We constructed a deep network containing a CNN part and a RNN part for the classification of EEG data. The architecture of our network is shown in Fig. \ref{network architecture}. The CNN part and the RNN part were combined through a reshaping operation. Firstly, EEG videos and optical flow were fed into the CNN part. Secondly, a reshaping operation merged and converted the outputs of the CNN part into a 2-dimensional feature vector. Then, the feature vector was fed into the RNN part with two recurrent layers. Finally, the outputs of the RNN part were fed into a dense layer with ReLU and a dense layer with softmax, to obtain a final category label. 
In our network, we apply 4*4 kernel for convolution layers and 3*3 kernel for max pooling layers. The recurrent layers contain 128 nodes and the full connection layer after the RNN unit contains 64 nodes.

There were two difficulties in training the network, including insufficient dataset and vanishing gradient problem in the time dimension while training the recurrent unit in RNN. 
Sufficient and balanced data are most important assumes in deep learning to satisfy the necessity of optimizing a tremendous number of weight parameters in neural networks. Unfortunately, this is usually not true for EEG signal because data acquisition is complex and expensive. However, EEG signal have a very high time resolution with current popular signal acquisition equipment. Herein, we train the CNN part with fully sampled video and use 12 frames of short video to train the RNN part. 
To against vanishing gradient problems while training the RNN part, replacing the simple perceptron hidden units with more complex units, such as Long-Short Term Memory (LSTM) \cite{Gers2000Learning} or Gated Recurrent Unit (GRU) \cite{Cho2014Learning} which function as memory cells, can help significantly.

\section{Experiments}

We implemented a mixed BCI-based rehabilitation support system for stroke patients with the EEG classification approach we proposed. 
Firstly, we obtained the image and depth of the operating platform by Microsoft Kinect2, and then applied a computer vision algorithm to identify targets and show them in the software interface. 
Then, choices were shown flickering in different frequencies, and the subjects utilized steady state visually evoked potential (SSVEP) to select one of them. Movement destination can be controlled by MI when the system is in move mode. Finally, the operation was performed by a robot arm with fingers.

\begin{figure}
    \centering
    \subfigure[Grasp]{
        \includegraphics[width=2.265in]{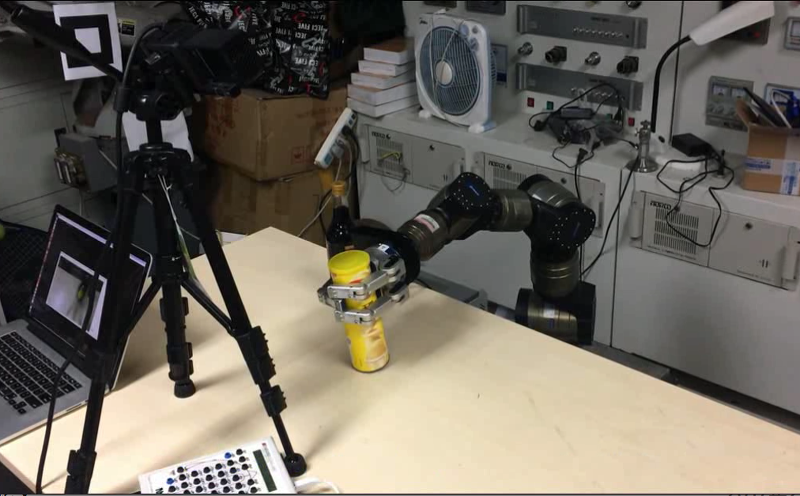}
        \label{figure:experiment:a}
    }
    \vspace{0in} \hspace{0in}
    \subfigure[Pour liquid]{
        \includegraphics[width=2.265in]{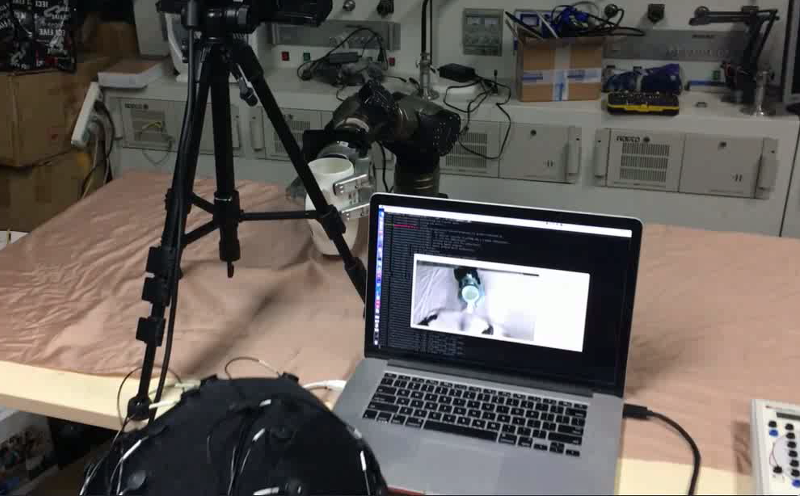}
        \label{figure:experiment:b}
    }
    \caption{Mixed BCI-based rehabilitation support system for stroke patients}
    \label{figure:experiment}
\end{figure}

With our rehabilitation support system, the subjects successfully performed some predefined operations through brain signals. 
In the grasp experiment (Fig. \ref{figure:experiment:a}), the subjects select a target, grasp it, move it to another position and put it down. In the pour liquid experiment (Fig. \ref{figure:experiment:b}), the subjects grasp a water cup, move it to the target position and pour it. These operations are critical for daily life, and can enhance the capacity for independent living of some special patients such as stroke patients.

\subsection{Dataset}

\begin{figure}[thpb]
    \centering
    \includegraphics[width=4.8in]{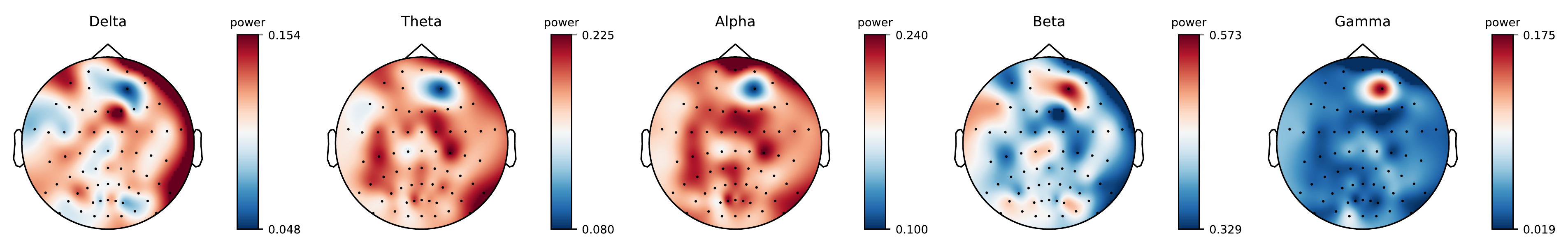}
    \caption{Visualization of power spectral density on our dataset}
    \label{psd_topomap}
\end{figure}

In the following analysis, we use the dataset collected by our system, from the MI data, while the four health subjects chooses the move direction in our software. The power spectral density after using five frequency filters is shown in Fig. \ref{psd_topomap}. They contain four categories (up, down, left and right imagined movements) signals for control movement direction, which are collected in 2s time-windows by 1000Hz sampling rate. Totally, we extracted dataset from 10 sessions, and used cross validation to distinguish training sets and test sets.

In addition, we apply our approach on the dataset \uppercase\expandafter{\romannumeral2}a from BCI competition \uppercase\expandafter{\romannumeral4}.
It contains EEG signal from nine subjects who perform four kinds of motor imagery (right hand, left hand, foot and tongue). These signals are recorded using 22 electrodes by 250Hz sampling rate and band-pass filtered between 0.5 and 100 Hz. For each subject, two sessions on different days were recorded and thus there are a total of 576 trials.

\subsection{Results}

\begin{figure}
    \centering
    \subfigure[Classification accurancy (\%) obtained from 10-flod cross validation]{
        \includegraphics[width=2.75in]{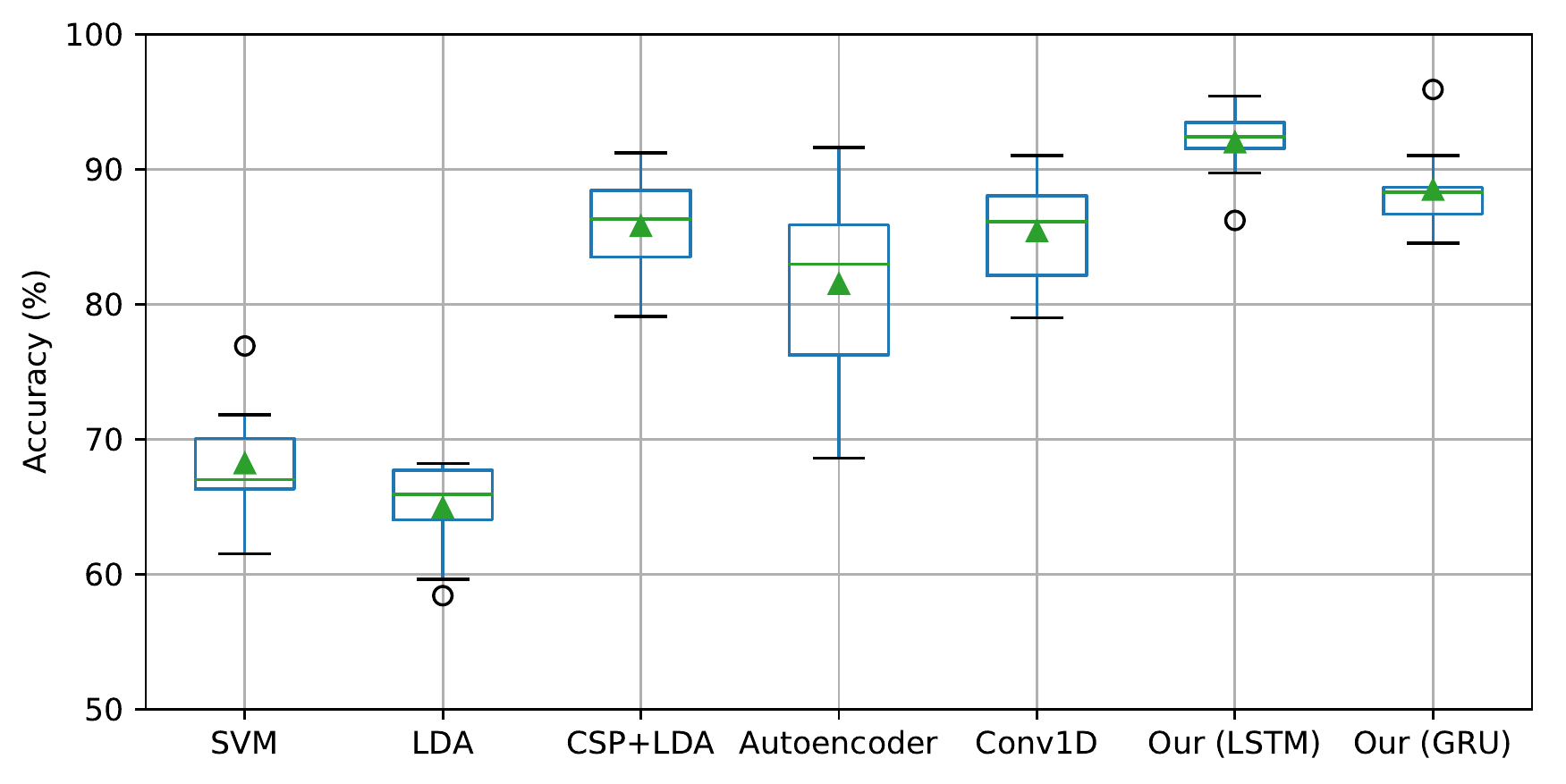}
        \label{figure:result:accrancy}
    }
    \vspace{0in} \hspace{0in}
    \subfigure[Accuracy of each epoch when training by our approach]{
        \includegraphics[width=1.79in]{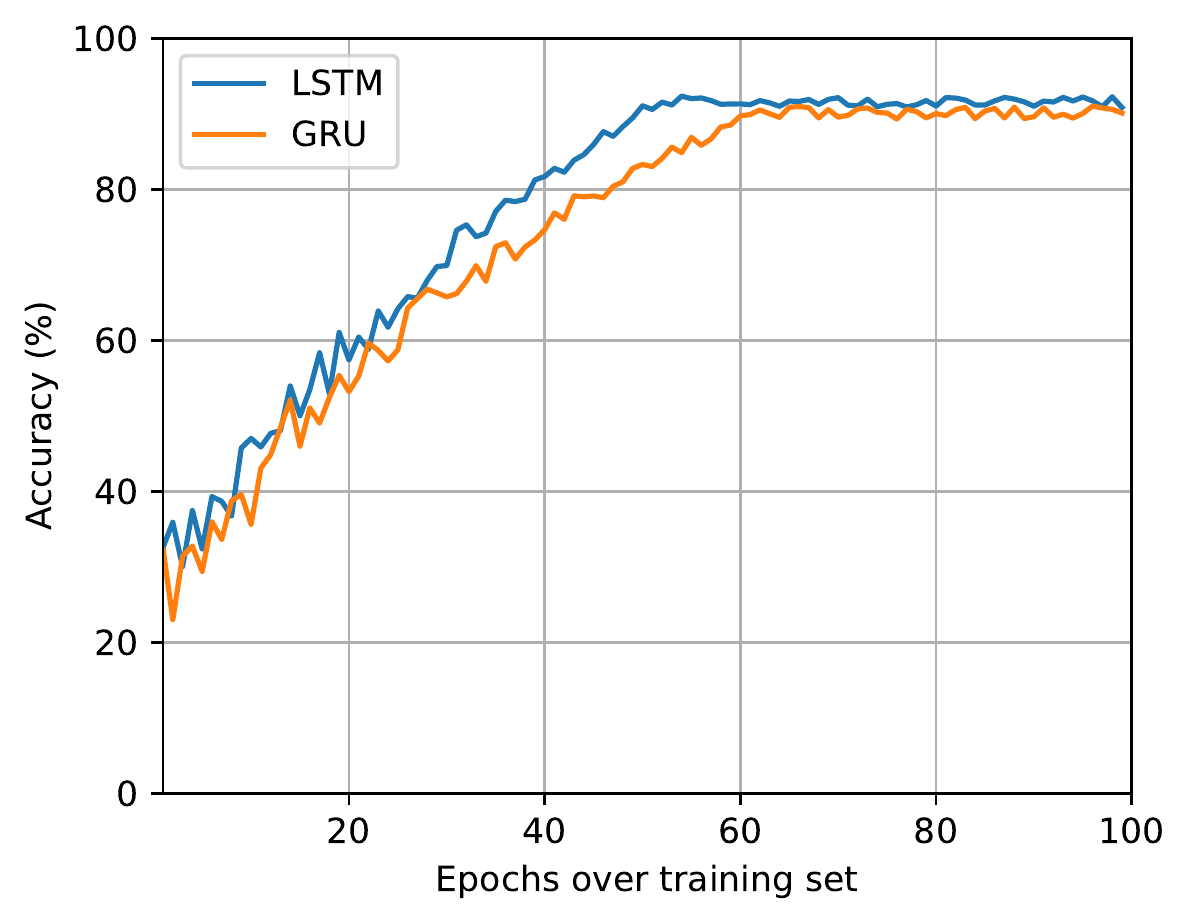}
        \label{figure:result:epochs}
    }
    \caption{Experiment results between our CNN-RNN network and other approaches based on the dataset collected from our rehabilitation support system.}
    \label{figure:result}
\end{figure}

We compared our approach against various classifiers commonly used in the field, including support vector machines (SVM), linear discriminant analysis (LDA), CSP+LDA, Autoencoder, Conv1D. 
SVM, LDA are the classic methods of machine learning. CSP is the most classical hand-designed feature and has been popular in this field for a long time. Autoencoder was introduced to this field recently. Conv1D is an intuitive attempt to apply CNNs to EEG classification. In our experiments, respectively, we tested the performance of these methods and our approach by applying LSTM or GRU as the basic elements of the RNN unit. 
We repeated many times by using every method we mentioned above, each time taking 9 sessions of data as training sets and 1 session of data as a test set. The performance results are shown in Fig. \ref{figure:result:accrancy} with offline training.
The experimental results show that our proposed approach can achieve more accuracy and stability, which is obviously superior to the traditional methods. There is no obvious difference between when we apply LSTM or GRU as the basic element of RNN, but it can reduce training time when applying GRU as the basic element of RNN. Moreover, it can be demonstrated that our approach can converge quickly and stably (Fig. \ref{figure:result:epochs}).

Furthermore, Table \ref{table: results comparison on bci competition dataset} presents the performance of our approach and traditional approaches on dataset \uppercase\expandafter{\romannumeral2}a from BCI competition \uppercase\expandafter{\romannumeral4}. It is clear that the our approach presented in this paper provides a significant improvement in classification accuracy over the traditional approaches. Results also suggested that our approach can achieve better performance when using LSTM.
These differences between LSTM and GRU can be due to the fact that LSTM has a more complex structure than GRU.

\begin{table}
\centering
\caption{Experiment results (\%) on dataset \uppercase\expandafter{\romannumeral2}a from BCI competition \uppercase\expandafter{\romannumeral4},
    $Sn$ is subject $n$ in the dataset.}
\label{table: results comparison on bci competition dataset}
\begin{tabular}{llllllllllll}
\hline\noalign{\smallskip}
~ & $S1$ & $S2$ & $S3$ & $S4$ & $S5$ & $S6$ & $S7$ & $S8$ & $S9$ & Avg & Std\\ \noalign{\smallskip}
\hline
\noalign{\smallskip}
SVM & 78.8 & 51.7 & 83.0 & 61.8 & 54.2 & 39.2 & 83.0 & 82.6 & 66.7 & 66.78 & 15.25 \\
CSP+LDA & 78.1 & 44.4 & 81.9 & 59.0 & 39.6 & 50.0 & 80.9 & 68.4 & 77.1 & 64.38 & 15.62 \\
Conv1D & 78.8 & 53.1 & 82.6 & 60.4 & 59.0 & 43.8 & 82.6 & 83.3 & 81.2 & 69.42 & 14.45 \\
Our approach(LSTM) & 78.8 & 62.5 & 83.0 & 63.5 & 67.7 & 45.8 & 90.3 & 85.8 & 72.6 & 72.22 & 13.17 \\
Our approach(GRU) & 90.6 & 41.0 & 95.1 & 68.1 & 47.6 & 54.9 & 90.3 & 64.9 & 80.6 & 70.34 & 18.79 \\
\hline
\end{tabular}
\end{table}

Our approach achieves superior accuracy over the traditional methods. However, due to the complexity of the network, careful design and optimization is needed to obtain satisfactory results. Herein, the training time of our network is much longer than other traditional methods because of two-step training strategy, especially when apply LSTM as the RNN unit.

\section{Conclusions}

In this paper, we propose a novel EEG classification approach, and build a mixed BCI-based rehabilitation support system. This rehabilitation support system can help stroke patients achieve a level of independence. 
The EEG classification problem is reduced to a video classification problem by converting EEG signal to gray-scale EEG videos. 
Moreover, optical flow has been introduced into this field, which can characterize the variant of EEG signal in the temporal dimension. 
To utilize the multimodal information of EEG, we project the position of electrodes to preserve the spatial information, apply multiple frequency filters to represent the spectral information, and utilize the time sequences information of EEG videos and optical flow to represent temporal information. 
We have constructed a deep neural network designed for these EEG videos and optical flow, and have partially solved the problem of insufficient EEG datasets by training the network in two steps. 
In future, EEG classification may be improved by state-of-the-art approaches from image classification and video classification. Particularly, we will apply the trained networks from image classification and video classification by transfer learning to solve the problem of insufficient EEG dataset.

\subsubsection*{Acknowledgments.} This work was supported by the National Natural Fund: 91420302 and 91520201. Thanks to the contributors of the open source software used in our system.

\bibliographystyle{splncs04}
\bibliography{iconip_reference}

\end{document}